\documentclass[letterpaper]{article}

\usepackage{natbib,alifeconf}  

\usepackage{graphicx}
\usepackage{todonotes}
\usepackage[hidelinks]{hyperref}

\usepackage{glossaries}
\usepackage{soul}
\usepackage{relsize}

\include{glossary}

\title{Centralized and Decentralized Control in Modular Robots and Their Effect on Morphology}

\author{Mia-Katrin Kvalsund$^{1}$, Kyrre Glette$^{1,2}$ \and Frank Veenstra$^1$ \\
\mbox{}\\
$^1$Department of Informatics, University of Oslo, Norway \\
$^2$RITMO, University of Oslo, Norway \\
mia.kvalsund@gmail.com}

\begin{document}

\maketitle

\begin{abstract}
In Evolutionary Robotics, evolutionary algorithms are used to co-optimize morphology and control. However, co-optimizing leads to different challenges: How do you optimize a controller for a body that often changes its number of inputs and outputs? Researchers must then make some choice between centralized or decentralized control. In this article, we study the effects of centralized and decentralized controllers on modular robot performance and morphologies. This is done by implementing one centralized and two decentralized continuous time recurrent neural network controllers, as well as a sine wave controller for a baseline. We found that a decentralized approach that was more independent of morphology size performed significantly better than the other approaches. It also worked well in a larger variety of morphology sizes. In addition, we highlighted the difficulties of implementing centralized control for a changing morphology, and saw that our centralized controller struggled more with early convergence than the other approaches. Our findings indicate that duplicated decentralized networks are beneficial when evolving both the morphology and control of modular robots. Overall, if these findings translate to other robot systems, our results and issues encountered can help future researchers make a choice of control method when co-optimizing morphology and control. 

\end{abstract}

\section{Introduction}

\begin{figure}[t]
\begin{center}
\includegraphics[width=3.1in]{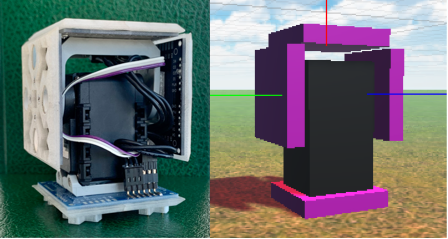}
\caption{\textbf{The EMeRGE module}. The left image is an example of a real EMeRGE module, and the right is our Unity simplification.}
\label{module}
\end{center}
\end{figure}

When co-optimizing morphology and control of modular robots, how do we optimize a controller for a robot that often changes its number of actuators and sensors? If a centralized approach is chosen, we must select a method to deal with disappearing actuators or the addition of new ones. Furthermore, although distributed control removes the issues of changing morphology, we must still then facilitate for global synchronization of the actuators. 
In this paper, we implement and discuss one centralized and three decentralized approaches to control and their effect on morphology. We suggest a decentralized control strategy that reuses control units across the robot body and demonstrate that such an approach leads to higher performance and more morphological development. 

Throughout Evolutionary Robotics' short history, there have been many approaches to co-optimizing morphology and control. Most notably, the work of Karl Sims showed virtual creatures evolved using a nested graph where body and control elements were connected \citep{Sims1994}. Controllers were duplicated as body parts were copied in a semi-distributed approach. \cite{Lipson2000} later showed a pipeline to transfer such virtual creatures to reality, where they used centralized control. Later, \cite{Cheney2013, Cheney2014} displayed soft-robots that evolved control through an indirect encoding using a Compositional Pattern Producing Network (CPPN). Similarly, \cite{auerbach2011evolving} used a CPPN encoding to generate a morphology and weights for a centralized continuous time recurrent neural network (CTRNN) controller. Both these approaches have reuse of control elements due to the indirect encodings.

\begin{figure*}[t]
\begin{center}
\includegraphics[width=5.8in]{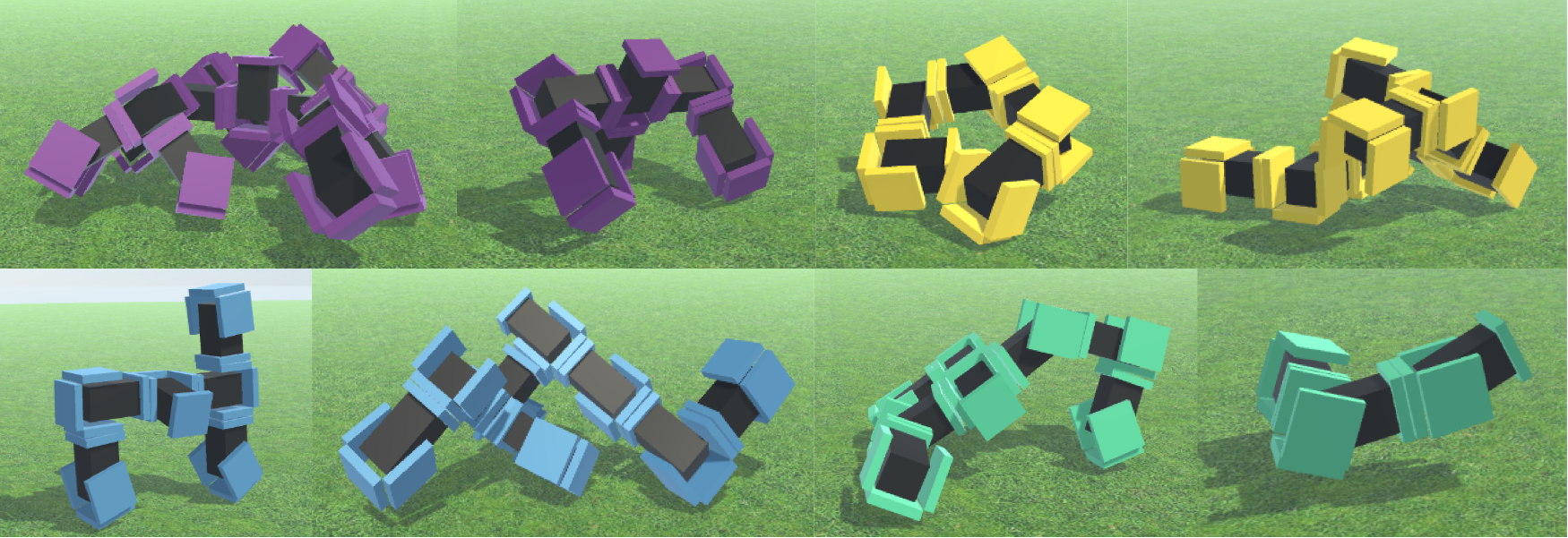}
\caption{\textbf{Examples of well-performing morphologies.} Purple and top left: Copy controller. Yellow and top right: Sine controller. Blue and bottom left: Decentralized CTRNN controller. Teal and bottom right: Centralized CTRNN controller.}
\label{robots}
\end{center}
\end{figure*}

A common problem when co-optimizing morphology and control is that of early convergence of morphology. As described by \cite{Joachimczak2016}, and later explored further by \cite{Cheney2016}, the morphology will reach its almost final form relatively early. Cheney et al. theorize that because the controller interacts with the environment through the interface of the body, changes to the body will scramble the control. However, this effect has not been studied much in modular robots. Decentralized control in modular robots could potentially decrease this issue because adding a new module with the same controller as its parent could still work without scrambling the overall performance.

Modular robotics (MR) concerns robots built from separable modules or units that encapsulate some function of a robotic system \citep{stoy2010self}. This is as opposed to an integrated design with no clear sectional modularity. The modules contain actuation, computation, energy, and sensing as needed, as well as some mechanism to connect and transmit to other modules. They can easily be reconfigured by hand or by machine, making them highly suited for rapid prototyping of robot designs. 

The field of modular robotics is quite young, with some of the first notable papers, like the CEBOT \citep{KawauchiY1992ASOS} and Fracta \citep{murata1994self} papers, being published in the 90s. Modular robots promised easily reconfigurable robots that could adapt their form to any use \citep{Yim2000}. Early systems such as the M-TRAN showed self-reconfiguration into shapes for walking and climbing \citep{Murata2002}, and \cite{zykov2005self} showed the first minimal example of self-reproduction in modular robots. Throughout the 2000s and early 2010s, the focus was mostly on exploring the promise of reconfiguration and creating novel mechanical solutions for the modules. 

Later \cite{marbach2004co}, inspired by the works of \cite{Sims1994} and \cite{Lipson2000}, started to co-evolve configuration and control. With works like Marbach  and  Ijspeert's Adam and EDHMoR \citep{Faina2013}, researchers started to experiment with the pipeline to create modular robots for any task. Evolutionary algorithms were uniquely suited for optimizing the robots, because the complexity of control scales exponentially with the number of modules \citep{marbach2004co}. Additionally, by co-evolving we avoid the limitations and biases a human designer would bring, hopefully producing more novel and better adapted solutions \citep{Faina2013}. 

Many systems in MR use sine wave generator controllers when the focus of the research is elsewhere \citep{Faina2013, Liu2017, Veenstra2017}, because they produce periodical movement with few parameters to optimize. This controller is a good baseline for the behavior of other controllers, as it is what we minimally expect from a decentralized controller. 

\begin{figure*}[t]
\begin{center}
\includegraphics[width=6.1in]{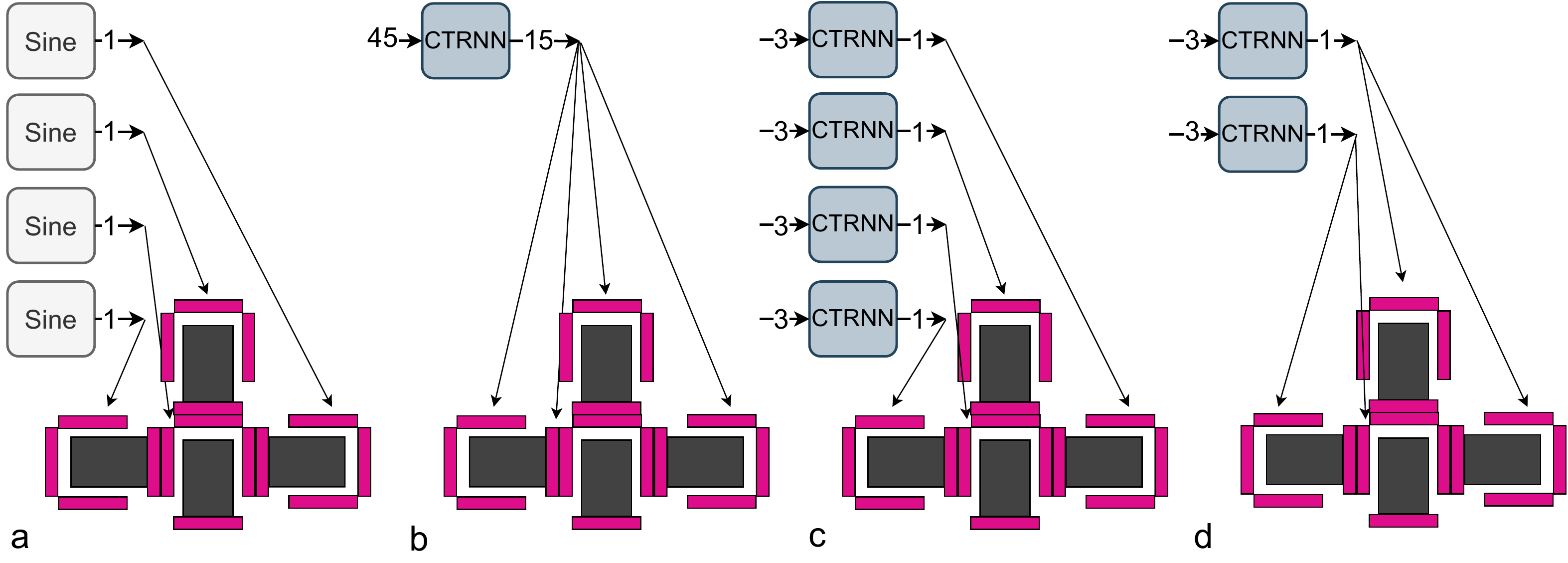}
\caption{\textbf{The controllers and how they could map to the modules in a small modular robot.} The arrows with numbers represent inputs (left side of boxes) and outputs (right side of boxes). a) The sine controller, b) The centralized CTRNN controller, c) The decentralized CTRNN controller, d) The copy controller.}
\label{controllers}
\end{center}
\end{figure*}

Evolved centralized control is something that is not much used in MR, as the focus early on was on hand-crafted centralized reconfiguration systems \citep{Murata2002}. Later works have also shown centralized control that focus on task execution and locomotion \citep{brunete2012behaviour}, however these are not evolved and do not necessarily scale well. It is still thought that some form of centralized control could be more suited to task execution \citep{Seo2019}, and so evolving centralized control should be explored.

Even so, decentralized control has also shown impressive results in task execution. \cite{christensen2006evolution} showed an example of a decentralized neural network controller, which could self-reconfigure and self-repair. Their modules' swarm-like, imperfect behavior was able to control above 3000 modules in simulation, showing the scalability of good decentralized control. Another good example of neural network control is \citeauthor{Jelisavcic2019}'s (\citeyear{Jelisavcic2019}) use of CPGs in the RoboGen modules. While this still results in distributed control, a CPPN encoding determines the CPG weights and can still enable module synchronization.

Our contribution to the field is two-fold: We present an investigation into centralized and decentralized controllers, and present a decentralized controller that reuses control units across the body. The controllers were implemented on a chain-type modular robot system.
Through measuring performance and morphological diversity, we evaluate which control approach is better suited for co-optimizing morphology and control in light of premature convergence of morphology. Our findings show that the controller that reuses control units leads to an increase in performance and morphological development, which advocates for reuse of control elements in controllers in general.


\section{Methods}

Our system consists of modular robots simulated in a flat ground environment and being measured on the task of locomotion. To co-optimize the morphology and controller, an evolutionary algorithm is used. This, as well as the four controllers we are investigating, will be presented below. 

For this project, an environment and simulated modular robots were built in Unity with the framework of ML-Agents
\footnote{Source code at \url{https://github.com/mia-katrin/Modbots}}. 
ML-Agents version 1.0.7 was used. Unity uses the Nvidia PhysX physics engine, which supports rigid body dynamics and updates physics steps every 0.02 seconds. The default physics engine settings were used. 

\subsection{The Modules}

We are using the EMeRGE module \citep{moreno2017emerge}, see \autoref{module}. It is a simple module with one servo motor, so that each module works like a hinge. It has four connection faces, one male at the base, and three female on the top and sides. The male connection face can only connect to the female ones and vice versa, meaning the robot will have a root module with three possible child modules, growing outwards in a tree-like structure. We do not allow connections to break. The joint is driven by a spring joint with 200 in spring torque and a damping value of 5. The max force is not constrained. Example morphologies can be seen in \autoref{robots}.

The original design for the modules includes infrared proximity sensors on each face. This was also implemented in our abstraction of the module, although the workings of the exact sensor model were not replicated. Instead, sensors register all distances through a ray cast, meaning the distances it can register are not capped and could get very high. For not registering a distance, for example from being angled towards the sky, a sensor returns -1. When a child module occupies a site, the sensor will register the small distance towards the child.

\subsection{The Controllers}

There are four controllers implemented in this system (\autoref{controllers}), each described below. For all of them, the output produced is directly the desired angle of a module's servo. The simulated module constrains the movement to +/-90$^{\circ}$.

Three of the controllers use a continuous time recurrent neural network (CTRNN)\footnote{The CTRNN implementation used is from the neat-python library \url{https://neat-python.readthedocs.io/en/latest/}}. 
This network was chosen to have the possibility of dynamic temporal behavior \citep{beer1995dynamics}. Here each node updates based on a differential equation, with neuron potentials as dependent variables. 

The CTRNN mutation operators can adjust weights and biases, change activation and aggregation functions, and cut away/disable or add/enable connections and nodes. The gene mutation rates are equal for all our CTRNN controllers and are given in the source code. These rates were then scaled by the controller's control mutation rate.

\subsubsection{Open-loop sine wave generator}

The open-loop sine wave generator is a decentralized controller that performs well because it produces periodic movement through sine waves, although it has no sensor input. In our case, the sine wave controller is used to provide a baseline to compare the other controllers to.
The controller is given by the function \begin{eqnarray}
y(t) = A * \sin ( w * t + p) + o\;
\end{eqnarray}
where $A$ is the amplitude, $w$ is the frequency, $t$ is time, $p$ is phase, and $o$ is offset. $y(t)$ is the controller output at time $t$ that is directly fed to the servo's desired angle. 

To enable synchronization between the modules, the frequency was fixed in all sine wave controllers and was not allowed to mutate. We noticed that without this, the sine controller was susceptible to choose local optima solutions. The sine wave controller has 3 parameters for each joint, meaning an average robot of 6 modules will have 18 parameters to optimize for the controller. When a module is added to the morphology,
it is instantiated with the control parameters of the parent module.

\subsubsection{Centralized CTRNN}

A straightforward approach to using a CTRNN for a modular robot is to simply gather all sensor outputs and feed them into one big CTRNN, that then outputs all controller actions. This leads to a fixed size CTRNN controller.  

In initial experiments with the centralized CTRNN controller, we tested numbers of inputs and outputs corresponding to 50 modules. Because there was a clear tendency to have a small number of modules, we tested with fewer inputs and outputs until finally we chose 15 to be the number of modules that would be controlled. This allowed the network to be as small as possible while still accommodating larger creatures at initialization, however it was very rare to see larger creatures with this controller. When a modular robot is smaller than 15 modules, the rest of the inputs to the network is set to 0.

The centralized CTRNN controller has 45 inputs and 15 outputs. It is initialized with 45 hidden nodes but is only 20\% connected. The order of mapping modules to input and output follows a depth first ordering of the modules, so that the first three inputs and first output goes to the root, the second three inputs and second output goes to its child, etc. 

The centralized CTRNN controller ends up having circa 600 connections and 60 nodes, each with respectively 3 and 5 parameters to tune, for a total of 2100 parameters.

\begin{figure}[t]
\begin{center}
\includegraphics[width=3.1in]{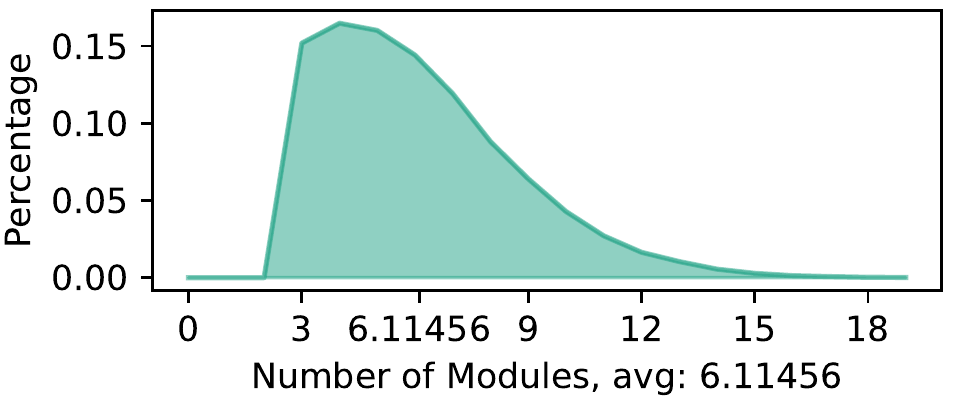}
\caption{\textbf{The distribution of number of modules in individuals in a random population before optimization.}}
\label{distribution}
\end{center}
\end{figure}

\subsubsection{Decentralized CTRNN}

Foregoing the benefits of a centralized brain, a decentralized approach leads to less parameters and a less complex optimization. The decentralized approach consists of each module getting its own CTRNN controller. When a module is added, it is instantiated with the control parameters of the parent module. 
Here, we used a small compact network of 3 inputs and 1 output, and with 3 hidden nodes and enabling all connections from the start, we get 4 nodes and 16 connections, totaling 68 parameters. For an average sized robot with one of these controllers in every module, this leads to about 400 parameters.  

\subsubsection{Copy Decentralized CTRNN}

The copy decentralized CTRNN controller, or the copy controller for short, is our alternative to the decentralized approach. It functions by having each robot keep a list of two CTRNN networks, which maps to different modules. The networks are the same as in the decentralized CTRNN controller. At initialization, these networks are clones, but as the optimization progresses, they will mutate separately. The modules will then mutate which network they use for control, theoretically allowing specialization. When a module is added to the morphology, it will use the same network as its parent module.

At the start of an evaluation, the networks are copied into their corresponding modules, hence the name. They then function independently of each other, and because of different sensor input such as detecting ground or the presence of child modules, they will likely behave differently. Nevertheless, it is reasonable to assume it will not achieve the level of specialization that the decentralized CTRNN controller can. While this might be a trade-off, we assume the copy controller will be quicker to achieve a good fitness, and possibly not be as dependent on number of modules. 

This controller, like the centralized CTRNN controller, is the same no matter the size of the robot. This gives us two controllers with 68 parameters, for a total of 136 parameters. Additionally, each module can change which controller they use, giving us a further average of 6 parameters. 

\subsection{Evolutionary Algorithm}

\begin{figure}[t]
\begin{center}
\includegraphics[width=3.5in]{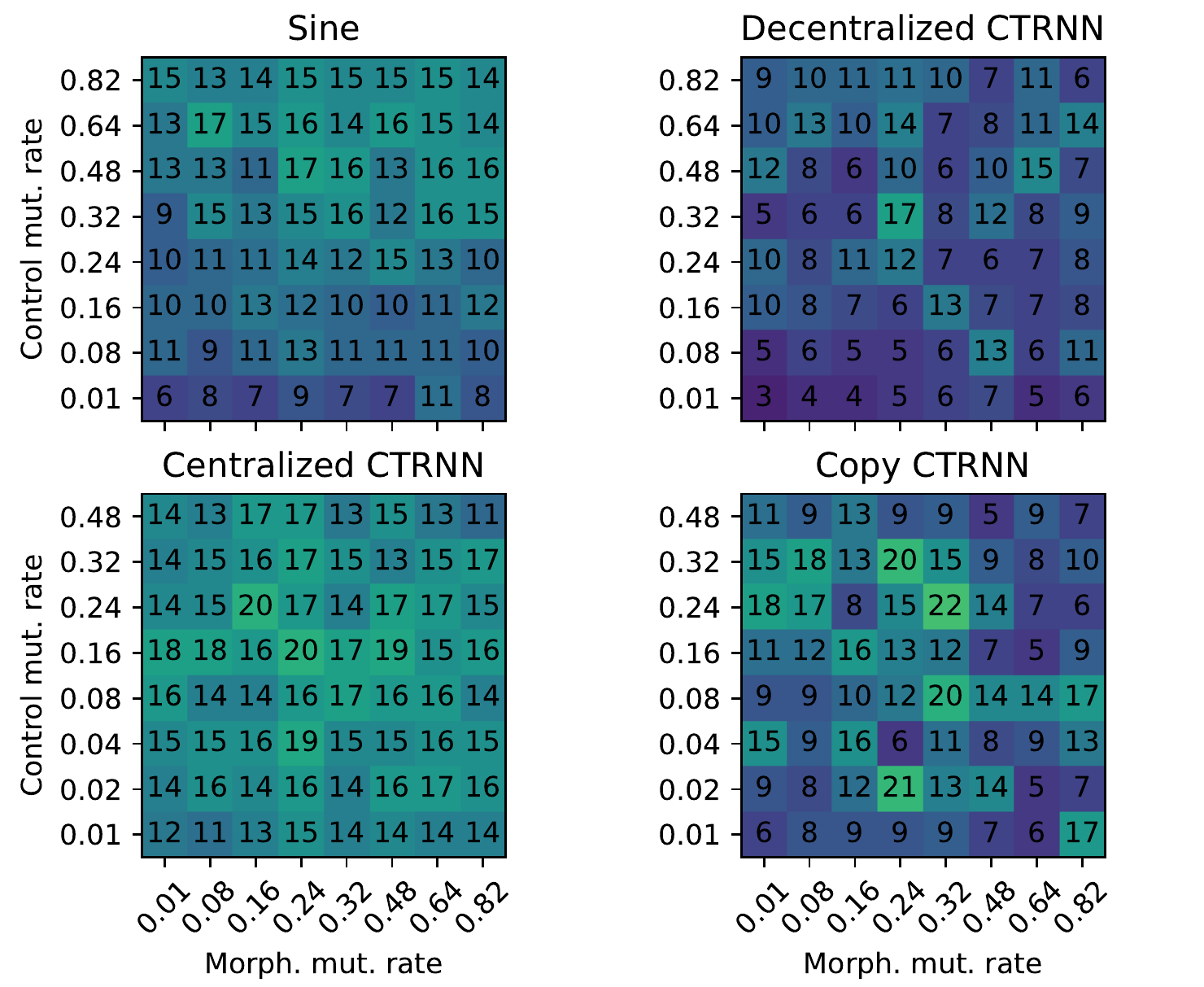}
\caption{\textbf{The grids filled for the parameter tuning}, values are rounded average fitness for each mutation rate pair. }
\label{grid_search}
\end{center}
\end{figure}

The evolutionary algorithm used had tournament selection, with a tournament size of 4, and generational replacement. It was implemented using the DEAP framework \citep{DEAP_JMLR2012}. Generational replacement was chosen because it can sometimes dislodge a population from early convergence. This happens because the best genotype will rarely be kept when the population is mutated and no elites are kept. Other parameters of the algorithm were also chosen to keep diversity. Most notably, the tournament selection size was small to increase selection pressure on the elites, while still being large enough to avoid a noisy evolutionary progression.   

The morphology and the controller had separate mutation rates. All controller gene values mutated with a Gaussian distribution based on the mutation power. For the rates, a parameter sweep was done for each controller on a grid of 8 values for both controller and morphology (\autoref{grid_search}), a total of 64 pairs. Each pair was run for 50 generations with a population size of 50, which totaled 2500 evaluations. This was done 4 times for each pair. Finally, the columns and rows in the grid were collapsed and plotted because there were a lot of variation between the pairs. The best performing column and row value were then chosen for each controller.

The 8 sweep values were chosen for each controller and the body based on whether they divided the mutation rate internally. In the body, the sine, and the decentralized CTRNN controller, the mutation rate is divided by the number of modules
so that creatures mutate the same amount no matter the size. This encourages the use of more modules because larger creatures are not more unstable solutions.
Since the average number of modules is 6 (\autoref{distribution}), the per-module mutation rate is 0.14 for the largest sweep value of 0.82. In the morphology mutation, this is further multiplied by circa 0.25 for each gene in the module. 

\begin{figure}[t]
\begin{center}
\includegraphics[width=3.1in]{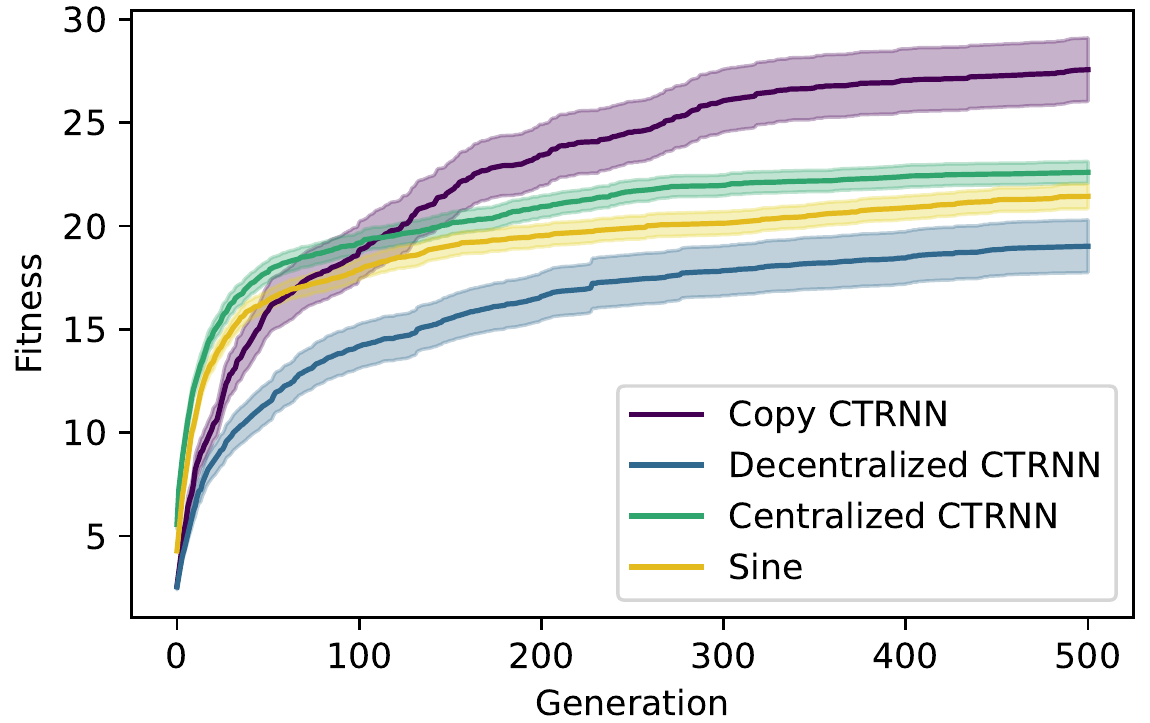}
\caption{\textbf{The fitness progressions for all four controllers.} The solid lines are averages of the best individuals in each generation, and the shaded areas represent the standard error.}
\label{fitness500}
\end{center}
\end{figure}

\subsection{Encoding of robots}

For the encoding of the robots, a direct encoding is used. A directed tree is generated from a root node, after which it is sent to the simulator. The simulator builds the robot and prunes any branches that collide, prioritizing keeping modules closer to the root. Modules that are not expressed are still kept in the genome.

The morphology can mutate by adding and removing modules, as well as changing the angle of modules. There is a slightly higher chance to get an add module mutation in order to bias creatures to grow. Additionally, one mutation will make a module duplicate one child branch to another connection site. This mutation was chosen to facilitate symmetry and larger jumps in the morphology landscape. 

\subsection{Fitness function}

The task that we measured the modular robots on was locomotion away from the origin during a set amount of time (100 steps of 0.2 seconds, roughly equaling 20 seconds in real time). Because this often leads to robots discovering the immediate optima of the somersault, or simply falling over, the robots were given 2 seconds to fall before the evaluation started. At the same time, the controller is not given input and will not give output. The fitness function is then
\begin{eqnarray}
Fitness = \sqrt{(x_{end} - x_{start})^2 + (y_{end} - y_{start})^2}\;
\end{eqnarray}
where $x_{start}$ is the x-position after 2 seconds of simulation, and $x_{end}$ is the x-position when the simulation ends. Likewise for y. The fitness is measured in units that correspond to the height of one module, which is circa 8 cm. A fitness of 30 would therefore mean 2.4 m has been travelled. The evaluation will stop early after 4 seconds from start if the robot has not moved in the last 2 seconds. 

\section{Results}

\subsection{Mutation rate sweep}

As can be seen in \autoref{grid_search}, the results from the sweep were often quite even. Only the copy and decentralized CTRNN controllers saw huge differences between different pairs, but all controllers had only minor differences after collapsing the data. Therefore, we settled on choosing a few of the ones that were contenders after the initial sweep and run a few more evolutionary runs on those. A winner would then often be clearer. The final parameters for all controllers can be seen in \autoref{rates}.

\begin{table}[ht]
\center{
\caption{\textbf{The mutation rate parameters chosen after the sweep.} Note that for the sine wave and decentralized CTRNN controller, as well as the morphology, the working mutation rate on one module is divided by the number of modules in the robot.}
\label{rates}
\vskip 0.25cm
\begin{tabular}{|c|c|c|}\hline
Controller & Morph. rate & Controller rate\\ \hline\hline
Sine wave & 0.32 & 0.64 \\
Centralized CTRNN & 0.24 & 0.16 \\
Decentralized CTRNN & 0.24 & 0.48\\
Copy CTRNN & 0.32 & 0.08\\
\hline
\end{tabular}
}
\end{table}

\subsection{Controller performance}

\begin{figure}[t]
\begin{center}
\includegraphics[width=3.2in]{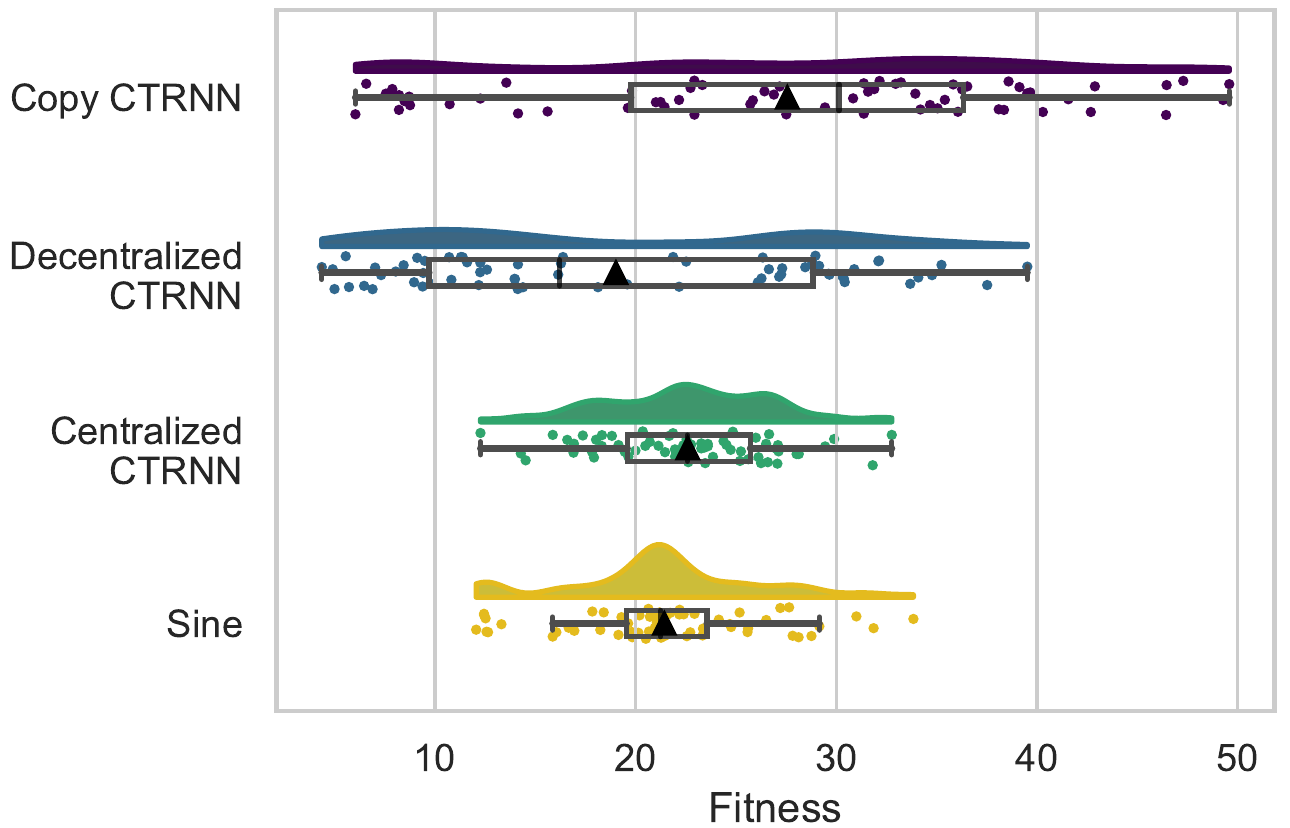}
\caption{\textbf{The final fitnesses for all runs for each controller.} The distributions are showed with the underlying points scattered below. A boxplot is placed over the scattered values, with the mean marked with a triangle.}
\label{raincloud}
\end{center}
\end{figure}

The final runs were done on populations of 50 individuals for 500 generations, for a total of 25 000 evaluations. This was done for all controllers 64 times, and the resulting performances can be seen in \autoref{fitness500}. \autoref{raincloud} shows the distribution of performances for the different controllers.

In order to get an overview of all significant differences, 6 two-sided Mann-Whitney U test were performed between all controllers at generations 50 and 500, a total of 12 tests. An alpha level of 0.05 was chosen. Because we were conducting multiple comparisons, Bonferroni correction was used. This gives us an adjusted alpha level of 0.05 / 12 = 0.00416.  

At generation 50, the sine and centralized CTRNN controllers were significantly different from the decentralized CTRNN controller (both p $<$ 0.0001). There was no significant difference between the sine and copy (p $>$ 0.2), sine and centralized CTRNN (p $>$ 0.07), copy and centralized CTRNN (p $>$ 0.06), and the copy and decentralized CTRNN controllers (p $>$ 0.01).

At generation 500, the copy controller was significantly different from the centralized CTRNN, the sine, and the decentralized CTRNN controllers (respective p-values p $<$ 0.003, p $<$ 0.0004, p $<$ 0.0002). There was no significant difference between the sine controller and the centralized and decentralized CTRNN controllers (both p $>$ 0.1), or between the centralized and the decentralized CTRNN controllers (p $>$ 0.04).

\subsection{Effect on morphology}

\begin{figure}[t]
\begin{center}
\includegraphics[width=3.in]{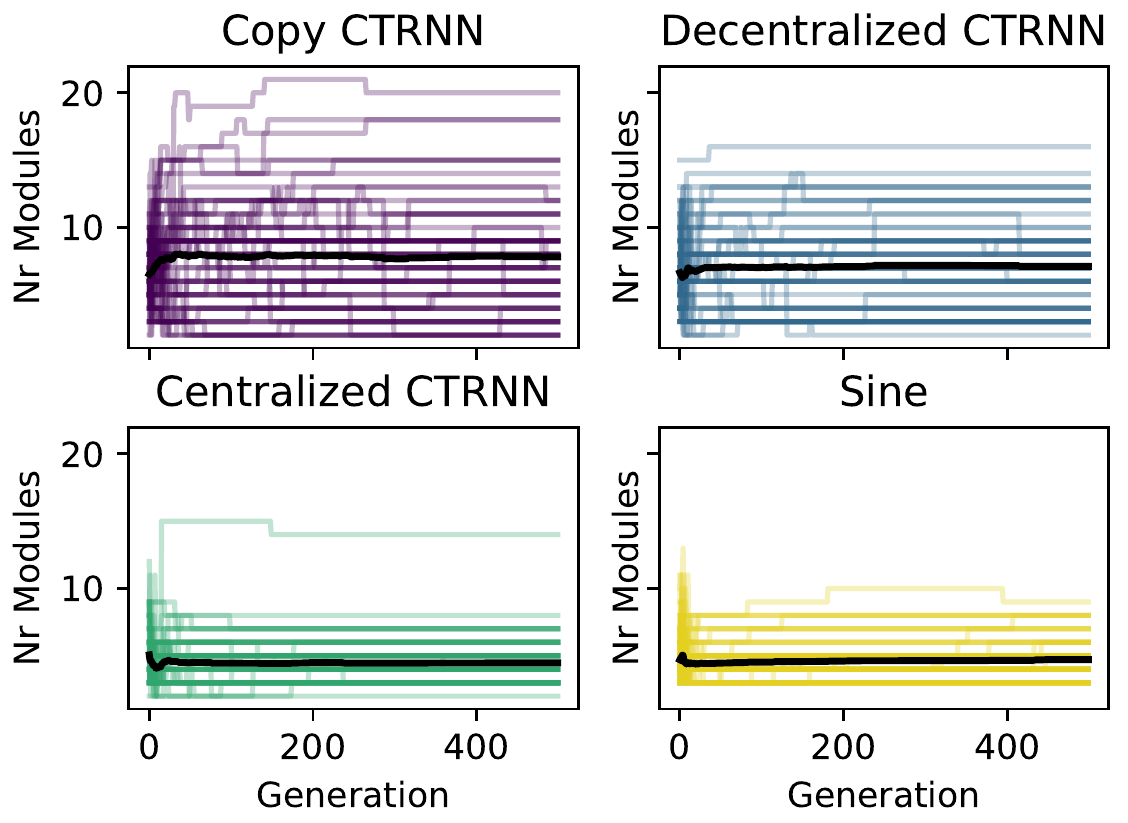}
\caption{\textbf{The progression of number of modules for all runs.} Only changes in number of modules that led to a more fit individual is shown.}
\label{nr_module_progression}
\end{center}
\end{figure}

In \autoref{nr_module_progression}, the progressions for number of modules in morphologies are plotted for all the controllers. Here, we can see that the sine and centralized CTRNN controllers both end up at a lower average number of modules than the copy and decentralized CTRNN controllers. To confirm if this was significant, as previously 6 two-sided Mann-Whitney U tests were performed with Bonferroni correction between the different count distributions. An alpha level of 0.05 was used, which means that with correction we consider p-values below 0.05 / 6 = 0.0083 as significant. 

Here we found that the copy and decentralized CTRNN controllers had no significant difference between them (p $>$ 0.4) and the sine and centralized CTRNN controllers likewise had no difference (p $>$ 0.2). However, the copy and decentralized CTRNN controllers were both different from the sine and centralized CTRNN controllers (all p $<$ 0.0001).  

Qualitatively, we recognize this from looking at the robots. The sine and centralized controllers had small, effective strategies while the copy and decentralized CTRNN controller both tended towards larger morphologies. The sine and centralized CTRNN controllers would do large, powerful movements, with some modules in the morphology not moving at all. As opposed to this, the copy and decentralized CTRNN controllers favored small, rapid movements. Here, the copy moved most its modules, while the decentralized CTRNN controller sometimes had unmoving modules. Example behaviors can be seen in the accompanying
video\footnote{\url{https://www.mn.uio.no/ifi/english/research/groups/robin/research-projects/cocomo/modbots.html}}.

\begin{figure}[t]
\begin{center}
\includegraphics[width=3.3in]{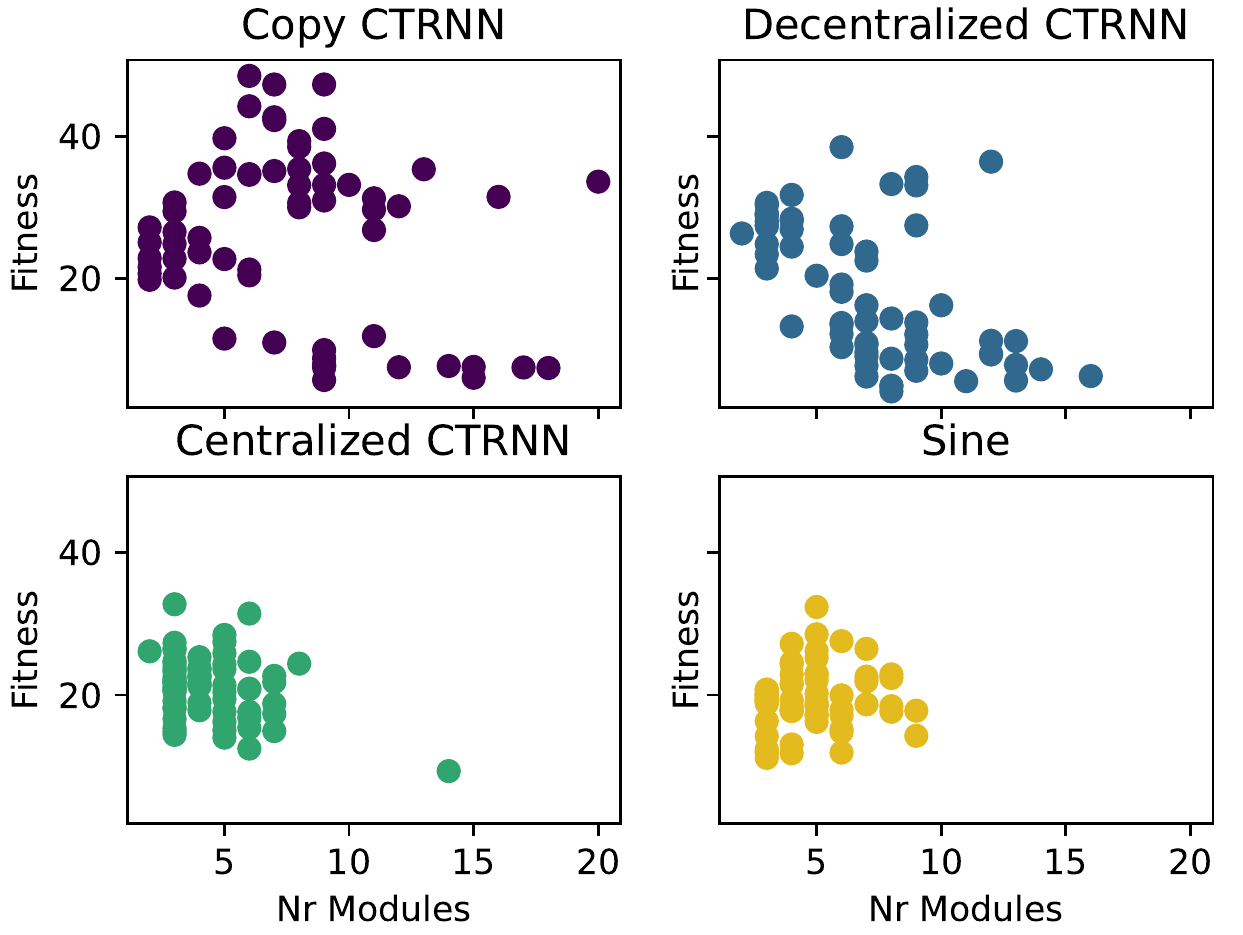}
\caption{\textbf{The expressed modules of all robots plotted against their fitness.}}
\label{correlation}
\end{center}
\end{figure}

\begin{figure}[t]
\begin{center}
\includegraphics[width=3.3in]{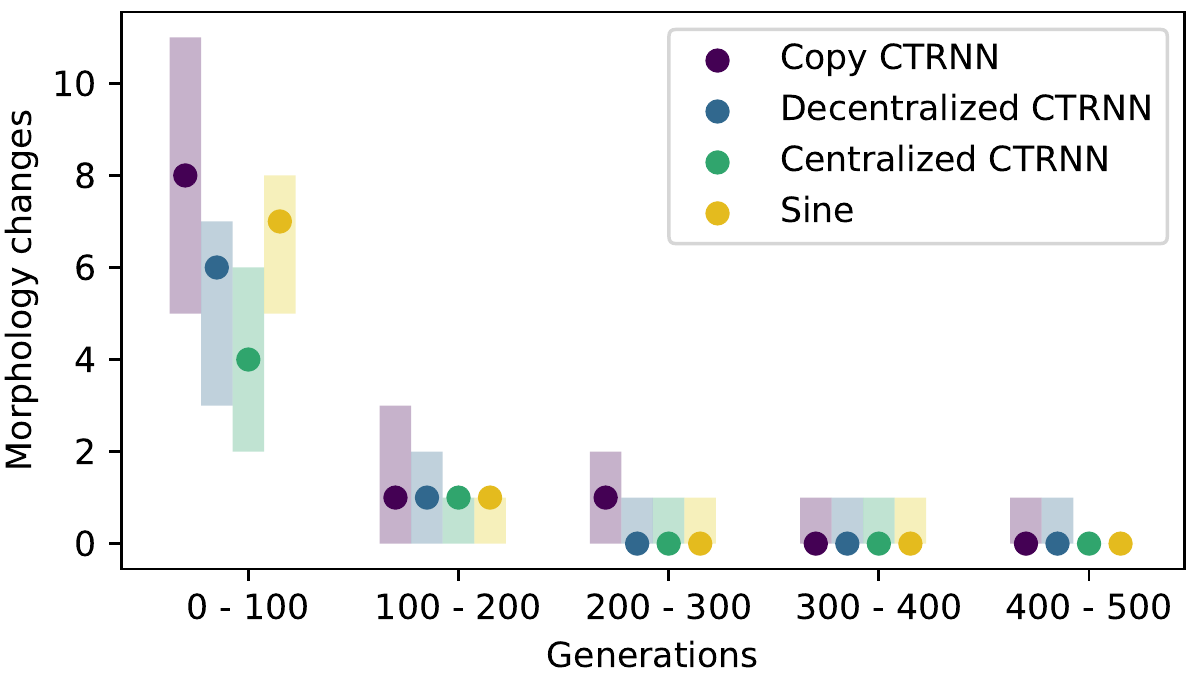}
\caption{\textbf{Number of morphology changes that led to better fit individuals in each generation interval}. The shaded areas are between the 25 and 75 percentiles, and the dots are the medians.}
\label{morph_changes}
\end{center}
\end{figure}

In \autoref{correlation}, we see that there seems to be a divide between larger creatures that get high fitness, and those that get low fitness. Presumably, this is because falling strategies tend to be larger in order to get further, but it is also interesting to see that some larger solutions did quite well. Especially the copy controller did well with larger morphologies, managing to produce a fitness on par with the centralized CTRNN and sine controllers with upwards of 10 expressed modules.

When looking into the issue of early convergence of morphology only, \autoref{morph_changes} was made. It shows shaded areas between the 25 and 75 percentiles of number of beneficial morphology changes in a run in each interval of 100 generations. It shows that the centralized CTRNN and sine controllers stop mutating morphology after 400 generations. The last two controllers kept mutating all the way up until the end. Interestingly, 72\% of the copy controller runs experienced morphology changes in the 100-200 generation interval, and more than 50\% had morphology changes in the 200-300 interval. 

\section{Discussion}

Our results have shown that the copy controller performs significantly better than other controllers when co-optimizing the morphology and control of modular robots. Since it duplicates behaviors, modules are more likely to be synchronized. Moreover, when a new module is added, a working control unit can be inherited that is already potentially useful. Even though the sine and decentralized CTRNN controllers had a similar feature of inheriting the parent module's control, the copy controller is more likely to be useful since it is already evolved to work in many different parts of the robot. In addition, because a control mutation affects multiple modules, it has an overall larger effect on the behavior of the robot compared to a mutation in the other controllers. Because of this, the copy CTRNN approach is less able to fine-tune a single controller compared to the other approaches and therefore may rely on morphological change to see a performance increase. This feature would thereby promote continued morphological diversification compared to the other controllers, as seen in \autoref{morph_changes}. 

From the fitness progressions, we can see that the sine and centralized CTRNN controllers converged rather fast compared to the other two. They also showed a pattern of quickly finding a final morphology of relatively small size, and then optimizing the controller. Meanwhile, the other two controllers spent time developing both and thus converged slower. Because having more modules means the robot has more potential force, allowing for more movement and higher fitness overall, the sine and centralized CTRNN controllers were then at a disadvantage. These results confirm that there is a trade-off between fine-tuning controllers and getting a good fitness with a small morphology while losing the potential of getting a higher fitness and a large morphology.

The distributions of solutions for the controllers vary wildly, as seen in \autoref{raincloud}. While the sine and centralized CTRNN controllers had a more solidly high performance, the decentralized CTRNN controllers both had very flat distributions, stretching from the worst to the best performances recorded. The lower fitnesses can be accounted for as robots that grow tall and fall in one direction, as some of these have been visually confirmed to be. The higher values of the copy, decentralized, and centralized CTRNN controllers often had rapid module movements that either led to small jumps or shuffling behaviors. Because of fixing the sine controller's frequency, this strategy was not available to the sine controller, and so its worse performance must at least be partially attributed to that. Even though it could have rivalled the others by growing larger, the CTRNN controllers' behavior was likely less complicated to evolve. Still, the sine and centralized CTRNN much more often arrived at very similar local optima, which tended to have the same fitness. Here, the centralized CTRNN controller had an advantage over the sine controller because it could optimize further by adding non-periodic movements. 

Because we had the same morphology mutation rate for the sine and copy controllers, we could expect similar morphological diversity from these. However, it is clear from our results that this is not the case. When keeping in mind that some of the more scalable strategies available to the CTRNN controllers were not available to the sine controller, it could simply be that there were less available good morphologies for the sine controller.

The centralized CTRNN approach that was implemented could evolve a network topology that connected to up to 15 modules. This means that evolving larger morphologies would involve the CTRNN accommodating for more outputs. Since the CTRNN in this case would then have even more parameters to optimize, we would expect the centralized CTRNN to converge even quicker. This could potentially be overcome by connecting parts of the neural network of the centralized CTRNN to the morphology and copying these parts of the CTRNN when a new module is introduced. Another possible cause for the rapid convergence seen in the centralized CTRNN is that the initial experiments to determine the supported number of modules only ran for 100 generations. Although we have no indication that it would, the centralized CTRNN approach could generate larger morphologies when given different mutation rate values.   

The copy and decentralized controllers both had issues of some number of unexpressed modules being added to the genome. These were unexpressed because they collided with other modules or the floor. 2 and 4 out of 64 samples from respectively the copy and decentralized CTRNN controllers had 10 to 20 unexpressed modules. Since they were unexpressed, the only effect they had on the individual was lowering the per-module mutation rate. This meant that the morphology in both, and the controllers in the decentralized CTRNN, would mutate less as the number of unexpressed modules grew. This bloating of the genome would in theory stabilize them from mutating. The two other controllers did not have issues with this.

Another issue is that of some pervasive local optima. Likely due to the angular shape of the EMeRGE modules, initial populations of robots found success with a single module dragging itself forward. In an attempt to avoid this, we constrained the robots to having a limp root module. This mitigated the problem somewhat, but similar strategies of using the corners of the female connection plates persisted all throughout the project. For example, the aforementioned jumping and shuffling behaviors are likely only possible because of the module shape. 

To avoid local optima, we tailored our algorithm to keep diversity, for example by having generational replacement. While using a diversity maintenance method could have minimized the occurrence of local optima solutions, it would have been difficult to parse which results were caused by the controllers and which were caused by the algorithm. Even so, in future work the same controller approaches could be tested with diversity enhancing methods to investigate what different control behaviors arise. 

Because the sensor implementation was not very realistic, it could be useful to focus on adding more realism and to measure the different controllers on tasks or environments that require more sensing. Here, we would be better able to study different strategies that emerged, and whether the controllers were as equally equipped for task execution as they were for locomotion. It seems probable that the centralized CTRNN would perform better than the others here. The decentralized approaches would benefit from communication between modules, and coupled CPGs such as the ones used by  \cite{ijspeert2007swimming} could likely work well here. 

Lastly, to test if these findings translate to other robotic systems, the same controller types should be implemented on a system with different modules and/or controllers. Since the EMeRGE module's female connection plates can be used for dragging and jumping, a less angular option like the RoboGrammar modules \citep{zhao2020robogrammar} could force the controllers to choose more complicated movements. Additionally, the previously discussed CPGs can be used to achieve periodic motion and would be a good option to further incorporate the sensors in locomotion. 

\section{Conclusion}

In this article we implemented and tested four controllers that were co-optimized along with a modular robot morphology. With testing three decentralized and one centralized controller, we got insight into how these can be done well and different challenges that arises for each approach. Markedly, we learned that there is significant advantage to simplify your controller to facilitate for global synchronization, as was found when the copy controller outcompeted the decentralized CTRNN controller. The copy approach allows for better new control of added modules, thus more morphological development, and larger jumps in the search space. Regarding centralized control, we highlighted the early convergence of morphology and performance when it comes to having a complex controller to optimize. Given that these findings translate to other controller networks and morphologies, they can aid future choices of control when co-optimizing morphology and control. 

\section{Acknowledgements}

This work was performed on the Fox resource, owned by the University of Oslo Center for Information Technology. This work was partially supported by the Research Council of Norway through its Centres of Excellence scheme, project number 262762.

\footnotesize
\bibliographystyle{apalike}
\bibliography{essay} 

\end{document}